\pdfoutput=1

\documentclass[11pt]{article}

\usepackage[preprint]{acl}

\usepackage{times}
\usepackage{latexsym}
\usepackage{lipsum}
\usepackage{xspace}
\usepackage{amsmath}
\usepackage{cleveref}
\usepackage{amssymb}

\crefname{section}{§}{§§}
\Crefname{section}{§}{§§}

\usepackage{amsthm}
\newtheorem{lemma}{Lemma}
\newtheorem*{lemma*}{Lemma}
\newtheorem{assumption}{Assumption}

\usepackage[T1]{fontenc}

\usepackage[utf8]{inputenc}

\usepackage{microtype}

\usepackage{inconsolata}
\usepackage{graphicx}
\usepackage{booktabs}
\usepackage{colortbl}
\usepackage{multirow}

\newcommand{\methodname}{{AdaEAGLE}\xspace}

\usepackage{ifthen}
\usepackage[normalem]{ulem}
\usepackage{CJKutf8}
\usepackage{algorithm}
\usepackage{algpseudocode}

\newcommand{\circleone}{\raisebox{-0.13\height}{\begin{CJK}{UTF8}{gbsn}①\end{CJK}}}
\newcommand{\circletwo}{\raisebox{-0.13\height}{\begin{CJK}{UTF8}{gbsn}②\end{CJK}}}
\newcommand{\circlethree}{\raisebox{-0.13\height}{\begin{CJK}{UTF8}{gbsn}③\end{CJK}}}
\newcommand{\circlefour}{\raisebox{-0.13\height}{\begin{CJK}{UTF8}{gbsn}④\end{CJK}}}

\title{\methodname: Optimizing Speculative Decoding via Explicit Modeling of Adaptive Draft Structures}

\author{Situo Zhang \and Hankun Wang \and Da Ma \and Zichen Zhu\\ {\bf Lu Chen\footnotemark[1] \and Kunyao Lan \and Kai Yu\footnotemark[1]}\\
X-LANCE Lab, Department of Computer Science and Engineering  \\
MoE Key Lab of Artificial Intelligence, SJTU AI Institute \\
Shanghai Jiao Tong University, Shanghai, China \\
\texttt{\{situozhang, wanghankun, chenlusz, kai.yu\}@sjtu.edu.cn} \\}

\begin{document}

\maketitle
\renewcommand{\thefootnote}{\fnsymbol{footnote}}
\footnotetext[1]{The corresponding authors are Lu Chen and Kai Yu.}
\renewcommand{\thefootnote}{\arabic{footnote}}

\begin{abstract}

Speculative Decoding (SD) is a popular lossless technique for accelerating the inference of Large Language Models (LLMs). We show that the decoding speed of SD frameworks with static draft structures can be significantly improved by incorporating context-aware adaptive draft structures. However, current studies on adaptive draft structures are limited by their performance, modeling approaches, and applicability. In this paper, we introduce \methodname, the first SD framework that explicitly models adaptive draft structures. \methodname leverages the Lightweight Draft Length Predictor (LDLP) module to explicitly predict the optimal number of draft tokens during inference to guide the draft model. It achieves comparable speedup results without manual thresholds and allows for deeper, more specialized optimizations. Moreover, together with threshold-based strategies, \methodname achieves a $1.62\times$ speedup over the vanilla AR decoding and outperforms fixed-length SotA baseline while maintaining output quality.

\end{abstract}

\section{Introduction}

\label{sec:intro}

Auto-regressive (AR) models are effective in language modeling but face latency issues due to their sequential token generation, particularly in Large Language Models (LLMs)~\cite{Brown2020GPT3, ouyang2022InstructGPT, Touvron2023LLaMA1, touvron_llama_2023}. Speculative decoding (SD) mitigates this by dividing decoding into two stages: a draft stage and a verification stage~\cite{chen_accelerating_2023, leviathan_fast_2023}. A smaller draft model predicts multiple tokens, which are verified by the larger model in one forward pass. By iterating these two stages, SD accelerates LLM while maintaining the original model's output distribution~\cite{chen_accelerating_2023}.

\begin{figure}
    \centering
    \includegraphics[width=\linewidth]{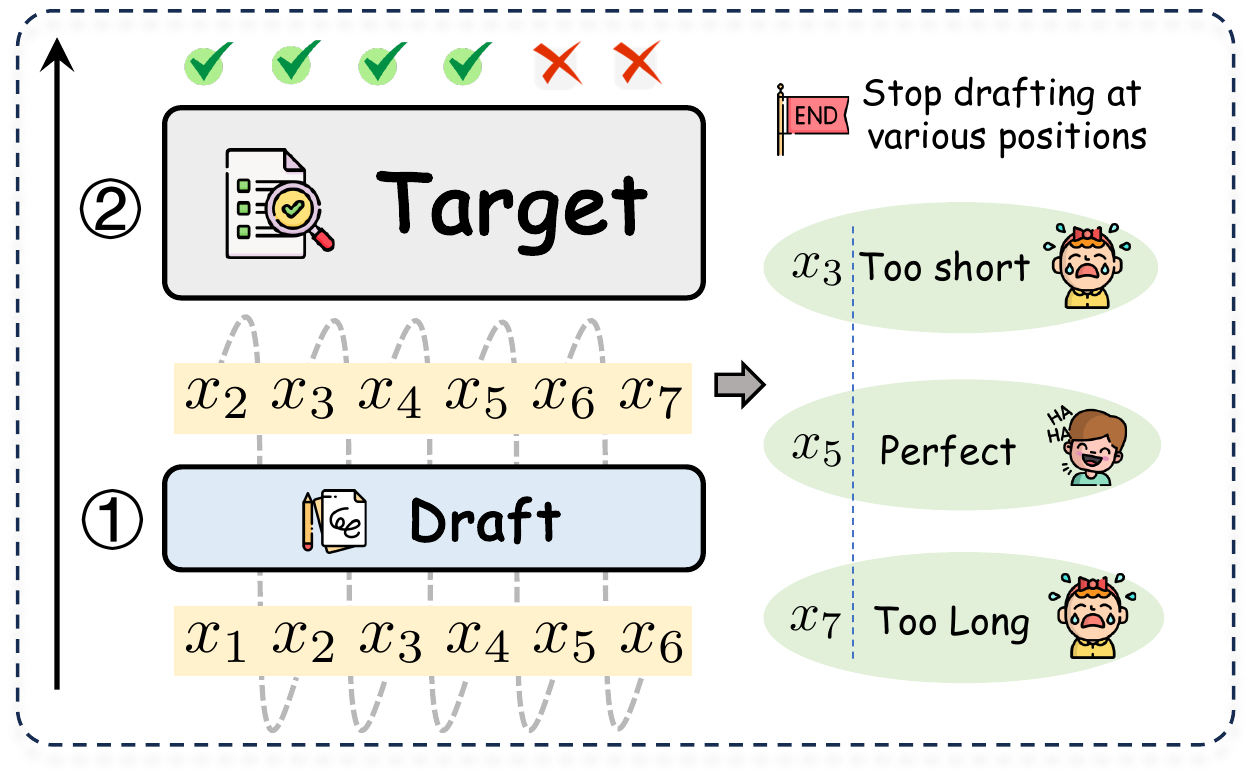}
    \caption{Motivation: finding the optimal draft stopping point to reduce target forward passes and minimize wasted computation.}
    \label{fig:motivation}
    \vspace{-1.5em}
\end{figure}

\begin{table*}[ht]
    \centering
    \small
    \begin{tabular}{lccc}
        \toprule
        \textbf{Method Name} & \textbf{Speculation Framework} & \textbf{Draft Structure Modeling} & \textbf{Manual Thresholds} \\
        \midrule
        \textbf{Draft \& Verify}~\cite{zhang_draft_2024} & Self-Speculative & Implicit & Required \\
        \textbf{SpecDec++}~\cite{huang_specdec_2024}     & Standalone Draft Model  & Implicit  & Required \\
        \textbf{DISCO}~\cite{mamou_dynamic_2024}         & Standalone Draft Model & Implicit & Required \\
        \textbf{PEARL} ~\cite{liu_parallel_2024}         & Standalone Draft Model  & None      & /      \\
        \textbf{EAGLE-2} ~\cite{li_eagle-2_2024}         & EAGLE~\cite{li_eagle_2024} & Implicit & Required \\  
        \textbf{DDD} ~\cite{brown_dynamicDDD_2024}       & EAGLE         & Implicit & Required \\
        \textbf{OPT-Tree}~\cite{wang_opt-tree_2024}      & EAGLE         & Implicit & Required \\
        \textbf{\methodname (Ours)}       & EAGLE        & Explicit & Not Required \\
        \bottomrule
    \end{tabular}
    \caption{Comparison of speculative decoding methods with adaptive draft lengths, in terms of base framework, draft boundary modeling, and manual threshold requirements.}
    \label{tab:spec_dec_comparison}
    \vspace{-1.5em}
\end{table*}

\begin{figure}
    \centering
    \includegraphics[width=0.8\linewidth]{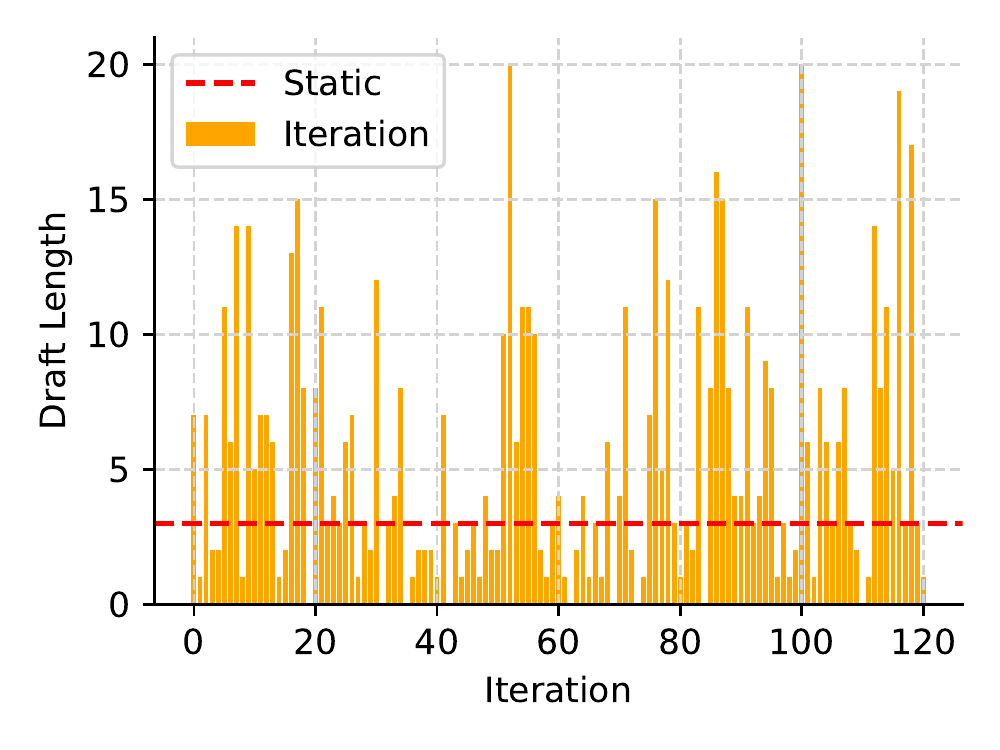}
    \caption{Maximum draft lengths accepted at different draft iterations via EAGLE.}
    \label{fig:iter_draft_length}
    \vspace{-1.5em}
\end{figure}

Most existing methods generate draft tokens using \textit{static} structures, such as fixed-length sequences~\cite{chen_accelerating_2023, leviathan_fast_2023} or static trees~\cite{cai_medusa_2024, li_eagle_2024}. However, studies have shown that the acceptance length of draft tokens strongly depends on the generated context~\cite{li_eagle-2_2024}. This implies that draft models with static draft structures cannot always exactly generate the optimal number of tokens, thus introducing more costly target model's forward passes and draft computation waste, as shown in Figure \ref{fig:motivation}. Figure~\ref{fig:iter_draft_length} further demonstrates the big difference in acceptance length of different draft iterations during the inference of a sample.
Therefore, to optimize decoding efficiency, an ideal draft model should use \textit{context-aware adaptive} draft structures, rather than using static structures. 
In fact, in our preliminary experiments (~\cref{sec:prel-discuss}), we further show that a draft length oracle (telling the draft model optimal draft length before generation) can improve the decoding throughput of fixed-length EAGLE \cite{li_eagle_2024} by 29\%. Thus, realizing adaptive draft structures is meaningful and rewarding.

Though multiple efforts have explored adaptive draft structures in different ways, as listed in Table \ref{tab:spec_dec_comparison}, they lack a comprehensive analysis and systematic approach. Their limitations can be summarized as: (1) \textbf{Unsatisfied performance}. Current methods are mostly based on inherent outputs and threshold control and fail to effectively model the optimal draft length. Experiments also show that there remains a significant optimization gap when compared to the topline performance of EAGLE-Oracle.
(2) \textbf{Hard to optimize}. The \textit{implicit} approach to modeling adaptive draft structures relies on the LM head output of the draft model, which is naturally trained on the original token sequences. This output cannot be further optimized for the specific goal of adaptive draft generation. Also, predefined fixed thresholds cannot fit all datasets and draft starting points, challenging the generalization capability. 
(3) \textbf{Obsolete applicability}. Earlier methods were not suitable for the current SotA framework, EAGLE~\cite{li_eagle_2024, zhou_survey_2024}. For example, SpecDec++~\cite{huang_specdec_2024} assumes draft generation follows a Markov Decision Process (MDP), which EAGLE does not conform. PEARL~\cite{liu_parallel_2024} is only applicable when the draft model’s inference cost per step is in the same order as the target model’s. (4) \textbf{Complex}. Methods requiring training involve intricate data construction and complex training processes~\cite{huang_specdec_2024, mamou_dynamic_2024}.

In this paper, we propose a novel method, \methodname, the first SD framework realizing explicit modeling of adaptive draft structures via a Lightweight Draft Length Prophet module (LDLP). The advantages of \methodname include:
(1) \textbf{Explicit modeling} of adaptive draft lengths. We use LDLP with simple inputs to predict the optimal draft length before generating the draft. Benefiting from this, LDLP neither relies on output logits nor needs manual set thresholds. (2) \textbf{Built on SotA} SD framework, EAGLE. Inspired by EAGLE and other works \cite{wu_language_2024}, we facilitate hidden states as well as output tokens to predict the draft length. (3) \textbf{Simple and easy}. Simplified data construction and highly parallelized training, making it easily integrated into existing SD frameworks.
Our results show that compared to vanilla AR decoding, \methodname achieves a speedup ratio of 1.61; it outperforms fixed-length EAGLE decoding with a speed improvement of 2\% and beats implicit-modeling methods (throughput 65.20 vs 64.43).

In Summary, the main contributions of our work are: (1) Provides a systematic analysis of the huge potential gain of realizing adaptive draft length. (2) Proposes a simple but effective SotA-based method, \methodname, to demonstrate that explicitly modeling of draft length is feasible and efficient in adaptive draft generation control. (3) Explicit modeling facilitates deeper and more delicate optimization and paves the way for improving adaptive draft structures in more complex scenarios, such as tree decoding and non-greedy decoding.

\begin{table*}[ht]
    \centering
    \small
    \begin{tabular}{lccccccccc}
        \toprule
        \textbf{Models} & \textbf{Draft Len.} & $\tau$ & \textbf{Tok/s} & $T_\textrm{total}$ & $T_\textrm{draft}$ & $T_\textrm{target}$ & $N_\textrm{draft}$ & $N_\textrm{target}$ & $N_\textrm{waste}$ \\
        \midrule
        Vanilla AR & - & 1.00 & 39.36 & - & - & - & - & - & -\\ \midrule
        \multirow{5}{*}{EAGLE~\cite{li_eagle_2024}}   & 2 & 2.35  &    58.89    & 692.93  & 60.60   & 576.93  & 36590  & 18295  & 11968 \\
           & 3 & 2.71  & 64.66 & 631.41  & 73.21   & 504.19  & 47535  & 15845  & 20459  \\
           & 4 & 2.95  & 63.84 & 638.77  & 85.32   & 499.93  & 58192  & 14548  & 29806  \\
           & 5 & 3.11  & 64.25 & 636.05  & 98.14   & 484.80  & 69035  & 13807  & 39896  \\
           & 6 & 3.22  & 61.98 & 656.57  & 110.89  & 492.21  & 80112  & 13352  & 50503  \\ \midrule
        EAGLE-Oracle & Dyn. & 3.41  & 83.46 & 483.64  & 48.74   & 383.66  & 33313  & 13952  & 0  \\
        \methodname  & Dyn. & 3.03 & 66.35 & 615.57  & 85.94   & 476.86  & 55649  & 14170  & 26871  \\
        \bottomrule
    \end{tabular}
    \caption{Performance comparison between the EAGLE model with various fixed draft lengths (ranging from 2 to 6) and adaptive-length draft models. $\tau$ and Tok/s are the acceptance length and throughput averaged over samples, respectively. $T_\textrm{total}$ represents the overall inference time, while $T_\textrm{draft}$ and $T_\textrm{target}$ represent the time spent on the draft model and target model, respectively. $N_\textrm{draft}$ and $N_\textrm{target}$ denote the number of tokens generated during the decoding process, and $N_\textrm{waste}$ indicates the number of draft tokens rejected by the target model during verification.}
    \label{tab:performance_metrics}
    \vspace{-1.5em}
\end{table*}

\section{Analysis of Adaptive Draft Length}
In this section, we provide a systematic analysis of the huge potential gain of integrating adaptive draft length control into SD. We start with introducing SD and its SotA variation, EAGLE~\cite{li_eagle_2024}. Then we analyze the optimal draft length under the EAGLE framework. Finally, based on pilot experiments, we discuss the significant benefit of adaptive drafts.
\subsection{Speculative Decoding}
Let \( T_{1:j}\) denote the sequence of tokens generated so far $t_1, t_{2}, \dots, t_j$ by an AR model (typically a LLM), and \( p(t_j) \) be the probability assigned to token \( t_j \) by the target model. Speculative decoding introduces a smaller draft model that guesses a sequence of candidate tokens \( \hat{T}_{j+1:j+k} \) along with their probabilities \( \hat{p}(\hat{t}_{j+i}) \) for \( i \in [1, k] \). 
The target model then computes the actual probabilities \( p(\hat{t}_{j+i}) \) for these tokens in parallel and decides the acceptance of the tokens in order. The acceptance of each draft token \( \hat{t}_{j+i} \) is determined by: $p_{\text{accept}}(\hat{t}_{j+i}) = \min\left(1, \frac{p(\hat{t}_{j+i})}{\hat{p}(\hat{t}_{j+i})}\right). $
If token \( \hat{t}_{j+i} \) is rejected, all successor tokens \( \hat{T}_{j+i+1:j+k} \) are discarded, and a new token is resampled from the distribution: $\operatorname{norm}\left(\max(0, p(\hat{t}_{j+i}) - \hat{p}(\hat{t}_{j+i}))\right).$ This process allows SD to validate multiple tokens in one pass, significantly reducing the number of sequential steps required in AR decoding. This method ensures that the output distribution of the SD is consistent with vanilla AR decoding of the target LLM~\cite{chen_accelerating_2023}. 

\subsection{EAGLE}
To improve draft quality, instead of training a separate small model, EAGLE (Extrapolation Algorithm for Greater Language Model Efficiency) reuses the target model’s input embedding and LM head, adding a trainable draft head in between~\cite{li_eagle_2024}. EAGLE’s first version used a static tree for draft validation in one forward pass, while the second version dynamically selected candidates based on confidence values~\cite{li_eagle-2_2024}. DDD~\cite{brown_dynamicDDD_2024} refined this method but still relied on confidence-based acceptance with manually set thresholds, as discussed in~\cref{sec:intro}.

\begin{figure*}[!ht]
    \centering
    \includegraphics[width=\linewidth]{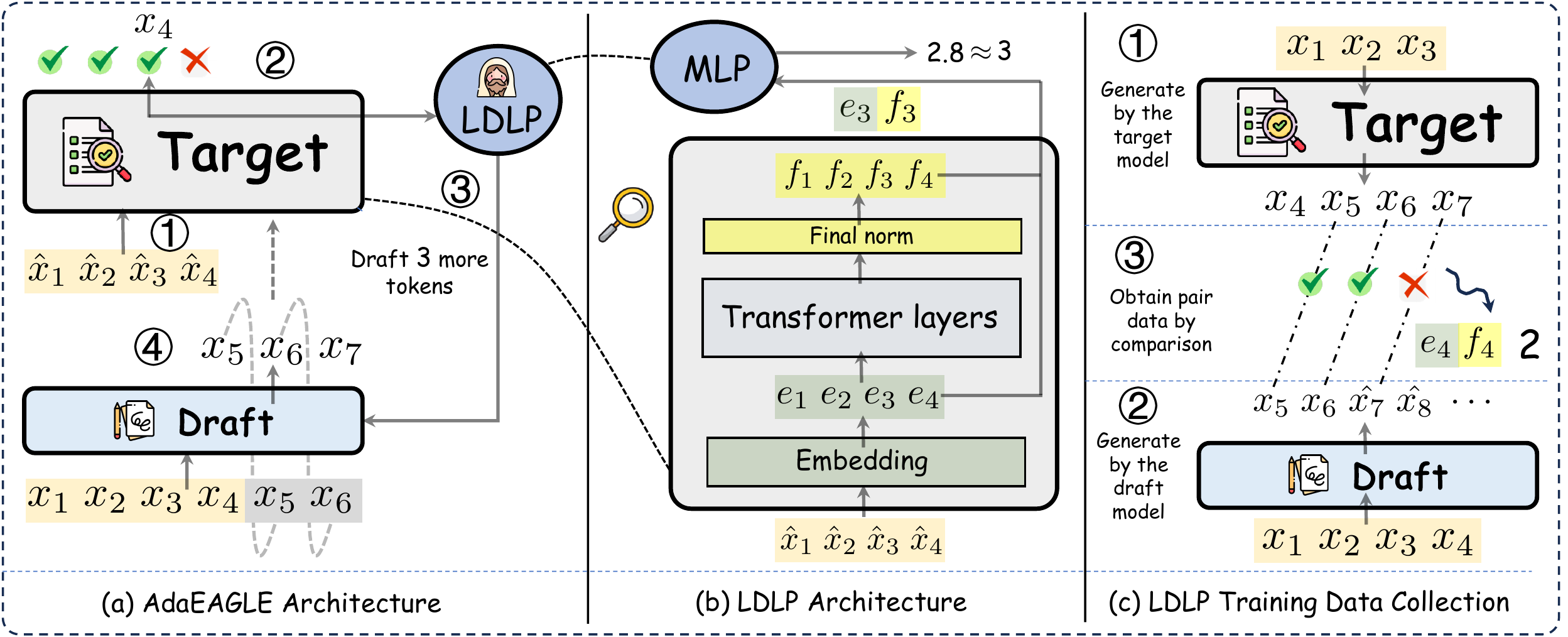}
    \caption{Illustration of \methodname framework. (a) The architecture of \methodname. The draft length are predicted by an LDLP module. (b) A closer look of LDLP. (c) An example of collecting paired training data of LDLP.}
    \label{fig:model-arch}
    \vspace{-1.2em}
\end{figure*}

\subsection{Draft-Length Oracle for EAGLE}
To facilitate the discussion, unless otherwise stated, the following paper is based on sequential draft generation and greedy decoding (sample temperature is set as $0$). We start by illustrating a new EAGLE-Oracle model with ideal adaptability via a draft-length oracle. The EAGLE-Oracle model uses the same architecture and parameters as the original EAGLE model, but during inference, a draft-length oracle module can tell the optimal draft length. 
In the context of greedy decoding, the optimal draft length is defined as the length that minimizes the number of forward passes required by the target model. In the context of greedy decoding and EAGLE, the optimal draft length is exactly the acceptance length (need an additional assumption, see the proof in Appendix \ref{apd-sec:oracle}).
By telling the draft model the optimal length before generating, EAGLE-Oracle ensures that the draft model outputs the necessary number of tokens, minimizing waste and maximizing efficiency.

\subsection{Benefit of Adaptive Draft Lengths}

\label{sec:prel-discuss}

By using the draft length oracle, EAGLE-Oracle achieves the theoretically optimal speedup under the given model parameters. We evaluated the acceleration effects of both fixed-length EAGLEs and EAGLE-Oracle. 
The experiment settings follow~\cref{sec:exp-setups}, with results shown in Table \ref{tab:performance_metrics}. The first five rows represent EAGLE with exactly 2 to 6 fixed-length draft token sequences, respectively, while the sixth row corresponds to EAGLE-Oracle. 
 
For EAGLE-Oracle, $N_\textrm{waste}=0$ since it generates exactly the optimal number of tokens at every draft starting position. As a result, the average acceptance length and the throughput of EAGLE-Oracle are much higher than those of EAGLE models with fixed-length drafts (improved by up to 42\%), demonstrating the huge potential reward of pursuing perfect adaptive draft lengths.

\section{\methodname}
In this section, we start by introducing the overall architecture of \methodname, the first SD framework that explicitly models the adaptive draft structure~(\cref{subsec:method_overall}). To achieve this, we carefully design a Lightweight Draft Length Predictor~(LDLP) which functions like a prophet by directly estimating the optimal draft length in advance during inference. More details of the LDLP are presented in \cref{subsec:method_ldlp}. Finally, we demonstrate how to train the LDLP~(\cref{subsec:method_train}).

\subsection{SD with Adaptive Draft Length}
\label{subsec:method_overall}
Denote the target model as $\mathcal{M}_T$ and the draft model as $\mathcal{M}_D$. The draft model is responsible for autoregressively generating drafts, while the target model verifies the correctness of all tokens in the draft in a single forward pass. The validated prefixes\footnote{A validated prefix is defined as the longest prefix in a draft that contains no incorrect tokens.} from the generated drafts are fed back into the draft model to continue decoding until the entire decoding process is complete.

Formally, for the interaction between $\mathcal{M}_T$ and $\mathcal{M}_D$ in the $r$-th iteration, let $T^r=\left(t_1,t_2,\ldots,t_j,\hat{t}_{j+1},\ldots,\hat{t}_{j+k}\right)$ denote the input sequence of $\mathcal{M}_T$, where $T_{1:j}^r$ is the generated formal sequence till last iteration, $T_{j+1:j+k}^r$ is the draft at current iteration, and $k\in\mathbb{N}^+$ is the draft length. The target model $\mathcal{M}_T$ verifies the draft via
\begin{equation}
    \label{eq:target}
\begin{aligned}
    k^\circ_r, t_{j+k^\circ_r+1} = \mathcal{M}_T\left(T^r, k\right),\\
    \bar{T}=T_{1:j+k^\circ_r}^r + \left(t_{j+k^\circ_r+1}, \right),
\end{aligned}
\end{equation}
where $k^\circ_r$ is the number of correct tokens in the draft and $t_{j+k^\circ_r+1}$ is an extra predicted bonus token during the verification~(Figure \ref{fig:model-arch}-(a)-\circleone\circletwo). Then, $\mathcal{M}_D$ generates the draft for the $(r+1)$-th iteration by
\begin{equation}
    \label{eq:draft}
    \begin{aligned}
        \left(\hat{t}_{j+k^\circ_r+2},\ldots,\hat{t}_{j+k^\circ_r+1+k}\right)=\mathcal{M}_D\left(\bar{T}, k\right),\\
        T^{r+1}=\bar{T}+\left(\hat{t}_{j+k^\circ_r+2},\ldots,\hat{t}_{j+k^\circ_r+1+k}\right),
    \end{aligned}
\end{equation}
where $+$ represents the concatenation operation~(Figure \ref{fig:model-arch}-(a)-\circlefour).

As we see in Equation \ref{eq:draft}, in standard speculative decoding, the length of the draft generated by $\mathcal{M}_D$ is a hyperparameter $k$ that is fixed in advance. According to the aforementioned analysis, during each iteration of interaction between $\mathcal{M}_T$ and $\mathcal{M}_D$, a significant gap between $k$ and the number of correct tokens in the draft $k^\circ_r$ can slow down the decoding speed of the model. Motivated by this, we design a a Lightweight Draft Length Predictor~(LDLP) which functions like a prophet by estimating the optimal draft length in advance~(Figure \ref{fig:model-arch}-(a)-\circlethree), i.e., predicting a value of $k_r\in\mathbb{N}^+$ closing to the optimal draft length $k^\circ_r$. Mathematically, the $k$ in Equation \ref{eq:target} and \ref{eq:draft} is replaced into $k_r$. 
We illustrate the details in Algorithm \ref{alg:overall}.

\subsection{Lightweight Draft Length Predictor}
\label{subsec:method_ldlp}

In designing LDLP, we follow the ``KISS'' (Keep It Simple and Stupid) principle. our goal is to ensure that its output closely approximates the number of correct tokens in the draft while keeping it as lightweight as possible. A heavy LDLP would increase computational overhead and data transfer between CPU and GPU, thereby negating the intended speedup of the speculative decoding. 

Based on this design principle, we implement a three-layer Multi-Layer Perceptron~(MLP) as the model structure for our LDLP. Prior studies on LLM future token prediction~\cite{pal_future_2023, hernandez_linearity_2024, wu_language_2024} point out that the hidden state of the last generated token contains the ``plan'' of generating subsequent future tokens. Inspired by this, for the $r$-th iteration, LDLP simply takes as input the embedding $e_{j+k^\circ_r}$ of the last token $\hat{t}_{j+k^\circ_r}$ in the validated prefix of the draft, along with its last hidden state $f_{j+k^\circ_r}$ after the final layer normalization in $\mathcal{M}_T$~(see Figure \ref{fig:model-arch}-(b)). Meanwhile, LDLP outputs a scalar in $\mathbb{R}$. Mathematically,
\begin{equation}
    \label{eq:LDLP}
    \begin{aligned}
        \bar{k}_{r+1}&=\text{Round}\left(\text{MLP}\left(e_{j+k^\circ_r}, f_{j+k^\circ_r}\right)\right),\\
        k_{r+1}&=\min\left\{k_\text{max}, \max\left\{0, \bar{k}_{r+1}\right\}\right\},
    \end{aligned}
\end{equation}
where the $k_\text{max}$ is a very slack hyperparameter that controls the upper bound of possible draft length.

\subsection{Training of LDLP}
\label{subsec:method_train}
The training of LDLP involves two main problems: training data collection and optimization criteria design.

\paragraph{Training data collection} As discussed in \cref{subsec:method_ldlp}, the training data for LDLP consists of pairs of $\left(\left[e_j;f_j\right], k^j\right)$, where $e_j$ and $f_j$ represent the embedding and last hidden state after final layer normalization in the target model for the $j$-th token, respectively. Here, $k^j$ denotes the optimal draft length starting from the $j$-th token. For a given prompt $T_\text{in}$ with $m$ tokens, we collect such pairs in the following three steps:
\begin{itemize}
    \item[1)] generate the output sequence $T^\text{out}$~($n$ tokens) by the target model $\mathcal{M}_T$ based on $T_\text{in}$~(Figure \ref{fig:model-arch}-(c)-\circleone)
    \item[2)] continue drafting $k_\text{max}$ tokens~(denoted as $\hat{T}^\text{out}$) by the draft model $\mathcal{M}_D$ based on $T_\text{in}+T^{\text{out}}_{:j}$~(for any $1\leq j \leq n$, see Figure \ref{fig:model-arch}-(c)-\circletwo)\footnote{The $k_\text{max}$ can be different from the value in Equation \ref{eq:LDLP}.}
    \item[3)] compare $\hat{T}^\text{out}$ to $T^\text{out}$ and calculate the length $k^j$ of the longest common prefix between them~(Figure\ref{fig:model-arch}-(c)-\circlethree)
\end{itemize}

\paragraph{Optimization criteria design} Let $k^\circ$ and $\bar{k}$ denote the ground truth and the prediction of LDLP, respectively. On the one hand, we desire $\bar{k}$ and $k^\circ$ to be as close as possible and we adopt $L_1$ loss to achieve this. On the other hand, we observe that the required time for a single forward pass of the target model is approximately $20$ times that of the draft model. Therefore, we hope that $\bar{k}$ is not less than $k^\circ$ as much as possible. To achieve this, we introduce a penalty coefficient $\lambda>1$ to scale up the $L_1$ loss when $\bar{k} < k^\circ$. Formally, the criteria is defined as
\begin{align}
    \mathcal{L} &= \left\{
    \begin{array}{ll}
    \lambda\cdot|\bar{k}-k^\circ| & \bar{k} < k^\circ \\
    |\bar{k}-k^\circ| & \text{otherwise}
    \end{array}
    \right.
    .
\end{align}

\begin{algorithm}[t]
\caption{Adaptive-Length Draft Decoding with LDLP}
\label{alg:overall}
\begin{algorithmic}[1]
\Require Prompt $T_\text{in}$
\Ensure Response $T_\text{out}$

\State $T\leftarrow T_\text{in}$ \Comment{$T^r$ in Equation \ref{eq:target}}
\State $k \leftarrow 0$ \Comment{the draft length $k_r$}
\While {not generating a terminator}
    \State Calculate $k^\circ_r$ and $\bar{T}$ \Comment{Equation \ref{eq:target}}
    \State Predict the draft length $k_{r+1}$ \Comment{Equation \ref{eq:LDLP}}
    \State Calculate $T^{r+1}$ \Comment{Equation \ref{eq:draft}}
    \State $k\leftarrow k_{r+1}$; $T\leftarrow T^{r+1}$
\EndWhile
\State $T_\text{out}\leftarrow \text{trim the prefix }T_\text{in} \text{ of } T$
\end{algorithmic}
\end{algorithm}

\subsection{Incorporate \methodname with Other Adaptive Draft Techniques}
Our explicit draft length modeling via LDLP can be easily combined with other approaches that utilize inherent model outputs (e.g., logits, entropy) for draft boundary modeling. By evaluating the per-step condition, the draft model can terminate generation based on both the predicted length and inherent threshold criteria. In experiments, we discuss the combination of \methodname and DDD \cite{brown_dynamicDDD_2024}, referred to as \methodname-DDD.

 \begin{table*}[!ht]
        \small
        \centering
        \setlength\tabcolsep{4pt}
        \small
        \resizebox{\textwidth}{!}{
        \begin{tabular}{lccccccccccccccc}
            \toprule
            \multirow{2.5}{*}{\textbf{Method}} & \multirow{2.5}{*}{\textbf{Draft Len.}} & \multicolumn{2}{c}{\textbf{MT-Bench}}& \multicolumn{2}{c}{\textbf{Alpaca}}& \multicolumn{2}{c}{\textbf{HumanEval}}& \multicolumn{2}{c}{\textbf{GSM8K}}& \multicolumn{2}{c}{\textbf{CNN/DM}}& \multicolumn{2}{c}{\textbf{Natural Ques.}} & \multicolumn{2}{c}{\textbf{Avg.}} \\
            \cmidrule(rl){3-4}\cmidrule(rl){5-6}\cmidrule(rl){7-8}\cmidrule(rl){9-10}\cmidrule(rl){11-12}\cmidrule(rl){13-14}\cmidrule(rl){15-16}
                           &  & \textbf{$\tau$} & \textbf{Tok/s}  & \textbf{$\tau$} & \textbf{Tok/s}  & \textbf{$\tau$} & \textbf{Tok/s}  & \textbf{$\tau$} & \textbf{Tok/s}  & \textbf{$\tau$} & \textbf{Tok/s}  & \textbf{$\tau$} & \textbf{Tok/s}  & \textbf{$\tau$} & \textbf{Tok/s}   \\
            \midrule
Vanilla AR & - & 1.00 & 	39.36 &  	1.00 & 	43.42 & 	1.00 & 	41.44 & 	1.00 	& 42.62 & 	1.00 & 	32.65 & 	1.00 	& 43.85 	& \cellcolor{blue!10}1.00  & \cellcolor{blue!10}40.56 \\ \midrule
            \multirow{5}{*}{EAGLE~\cite{li_eagle_2024}}
& 2 & 2.35 &	58.89 &	2.27 &	59.33 &	2.48 &	67.01 &	2.42 &	63.16 &	2.21 &	47.41 &	2.07 &	54.34 &	\cellcolor{blue!10}2.30 &	\cellcolor{blue!10}58.36 
\\       & 3 & 2.71                 & 64.66       & 2.61                 & 65.01       & 2.92                 & 75.50                & 2.80                 & 69.36                & 2.47                 & 50.32       & 2.29                 & \textbf{57.44}       & \cellcolor{blue!10}2.63 & \cellcolor{blue!10}63.72 \\
 & 4       & 2.95                 & 63.84                & 2.82                 & 62.94                & 3.24                 & 76.11                & 3.08                 & 69.25                & 2.61                 & 48.89                & 2.43                 & 55.25                & \cellcolor{blue!10}2.86 & \cellcolor{blue!10}62.71 \\
 & 5       & 3.11                 & 64.25                & 2.95                 & 62.88                & 3.44                 & 77.95       & 3.27                 & 69.83       & 2.69                 & 48.36                & 2.51                 & 54.00                & \cellcolor{blue!10}3.00 & \cellcolor{blue!10}62.88 \\
 & 6       & 3.22                 & 61.98                & 3.04                 & 60.18                & 3.57                 & 75.05                & 3.37                 & 66.77                & 2.73                 & 46.16                & 2.56                 & 51.45                & \cellcolor{blue!10}3.08 & \cellcolor{blue!10}60.27 \\ \midrule
DDD \cite{brown_dynamicDDD_2024} & Dyn. & 3.21                 & 66.33                & 3.02                 & 65.41                & 3.59                 & 77.71                & 3.41                 & 69.39                & 2.65                 & \textbf{50.53}                & 2.49                 & 57.18                & \cellcolor{blue!10}3.06 & \cellcolor{blue!10}64.43 \\ \midrule
\methodname  & Dyn.   & 3.03                 & 66.35                & 2.86                 & 64.67                & 3.41                 & 77.55                & 3.23                 & 70.68                & 2.60                 & 49.99                & 2.43                 & 57.41                & \cellcolor{blue!10}2.93 & \cellcolor{blue!10}64.44 \\
\methodname-DDD   & Dyn.  & 3.12                 & \textbf{66.86}                & 2.93                 & \textbf{66.16}                & 3.53                 & \textbf{79.07}                & 3.34                 & \textbf{71.20}                & 2.63                 & 50.52                & 2.45                 & 57.36                & \cellcolor{blue!10}3.00                 & \cellcolor{blue!10}\textbf{65.20} \\
            \bottomrule
        \end{tabular}}
        \caption{Average accpeted tokens $\tau$ and throughput \textbf{Tok/s} of different methods on six benchmarks. We include standard EAGLE speculative decoding method with fixed draft length ranging from 2 to 6 tokens. } %
        \label{tab:main-results}
        \vspace{-1.5em}
    \end{table*}

\section{Experiments}

\subsection{Experimental Setups}
\label{sec:exp-setups}
\paragraph{Model} We conduct the experiments with Vicuna-7B-v1.3~\cite{vicuna2023} and the paired EAGLE~\cite{li_eagle_2024} draft model. Our LDLP model is a three-layer MLP with residual connections, and the output hidden states are mapped to the length scalar through a linear transform.
\paragraph{Training} For a fair comparison, we use the ShareGPT dataset with 68000 dialog samples to train our length prediction model. We first ask Vicuna to auto-regressively generate responses and give the prompts for each dialog sample. Then we follow the data collection process detailed in~\cref{subsec:method_train}. For the training of LDLP, we keep the base target model and corresponding Eagle draft model frozen. We set the start learning rate to 5e-5 and use a cosine learning rate scheduler. LDLP is trained with batch size 128 for 5 epochs trained within 30 minutes on 8 A800 80GB GPUs, making \methodname training-efficient.
\paragraph{Benchmarks} Following the evaluation setup of EAGLE~\cite{li_eagle-2_2024}, we select six representative text generation benchmarks across diverse domains to evaluate our method. These benchmarks include the multi-turn conversation task MT-Bench~\cite{zheng_mtbench_2024}, the instruction-following task Alpaca~\cite{stanford_crfm}, the code generation task HumanEval~\cite{Chen2021HumanEval}, the mathematical reasoning task GSM8K~\cite{Cobbe2021TrainingVT}, the summarization task CNN/Daily Mail~\cite{nallapati-2016-abstractive}, and the question-answering task Natural Questions~\cite{kwiatkowski-2019-naturalques}. All experiments are conducted on a single A800-80G GPU with a batch size of 1, using greedy decoding (temperature set to 0.0) and FP32 precision.
\paragraph{Metrics} Since we do not modify the target model or the draft model, the generation quality remains consistent with EAGLE and prior speculative decoding methods that achieve lossless acceleration. Therefore, we do not report quality metrics for the benchmarks. Instead, we use the following two metrics to evaluate acceleration performance: (1) average acceptance length ($\tau$), defined as the number of accepted tokens per decoding step, and (2) throughput (Tok/s), defined as the number of tokens generated per second.

\subsection{Main Results}
\paragraph{Baselines} We select the original EAGLE method with a fixed draft length as our baseline. During each decoding step, EAGLE autoregressively generates a fixed number of draft tokens (ranging from 2 to 6). We also compare our method with the recently proposed adaptive draft length approach, DDD \cite{brown_dynamicDDD_2024}, applied to EAGLE. DDD is a threshold-based method with implicit draft structure modeling that controls the termination of draft generation. Specifically, DDD uses the probability of a draft sequence, which is calculated as the product of the probabilities of all tokens generated up to that point. If the product exceeds a given threshold, the draft model continues generation; otherwise, it terminates. We set the threshold to the value that performed best on MT-Bench and test it on other benchmarks.

\paragraph{\methodname-DDD} We also incorporate our draft length prediction method with the threshold-based method of DDD, resulting in \methodname-DDD. The draft model exit generation iff. the draft step exceeds the predicted length and the probability of the draft sequence is lower than the threshold.\\

Table~\ref{tab:main-results} compares our method with other baseline methods in terms of average accepted tokens and throughput (Tok/s). On average, \methodname achieves the highest throughput, outperforming all variations of the fixed draft length with the standard EAGLE decoding method. Fixed draft length methods can only achieve the best performance on specific benchmarks, for example, EAGLE with fixed draft length 3 achieves the best throughput on Natural Questions. This is because the response to this question-answering task is more dynamic and the draft model cannot effectively match the output of the target model, hence resulting in a short average accepted token. In this case, a short fixed length like three would save waste tokens. However on the code generation benchmark HumanEval, where the output is more predictable due to the structure of code and draft models have more acceptable tokens, a larger fixed draft length like five would reduce the number of target forward times. While \methodname with adaptive draft length demonstrates its adaptability across different tasks.

\begin{table}[t]
    \centering
    \small
    \begin{tabular}{lcccc}
        \toprule
        \multirow{2.5}{*}{\textbf{Method}} & \multicolumn{2}{c}{\textbf{MT-Bench}} & \multicolumn{2}{c}{\textbf{Alpaca}} \\
        \cmidrule(rl){2-3}\cmidrule(rl){4-5}
                                           & $\tau$ & \textbf{Toks/s}  & $\tau$ & \textbf{Toks/s} \\
        \midrule
        w/o Penalty                         & 2.81         & 64.62             & 2.62         & 63.08          \\
        w/ Cls                      & 2.87         & 64.27             & 2.72         & 63.64          \\
        \methodname                           & \textbf{3.03}         & \textbf{66.35}             & \textbf{2.86}         & \textbf{64.67}          \\
        \bottomrule
    \end{tabular}
    \caption{Ablation experiment results on MT-Bench and Alpaca about the training loss design. ''w/o Penalty" indicate that the model is trained without L1 loss penalty. ''w/ Cls" indicate that the model is trained with classification loss.}
    \label{tab:ablation-penalty}
\end{table}

Table~\ref{tab:performance_metrics} provides a breakdown analysis of the factors contributing to acceleration on MT-Bench, showing the inference time spent on the draft and target models for each draft setting, along with the number of tokens generated by each. For EAGLE with a fixed draft length, as the draft length increases, the number of tokens generated by the draft model ($N_\text{draft}$) grows from 47,525 to 80,112, resulting in fewer target model forward passes ($N_\text{target}$). However, this also leads to a surge in the number of wasted tokens ($N_\text{waste}$) due to the drastic fluctuation in maximum draft lengths at different positions, as shown in Figure~\ref{fig:iter_draft_length}. This trend is directly reflected in the time spent on each component: as $N_\text{draft}$ increases, $T_\text{draft}$ also increases, while $T_\text{target}$ decreases due to the reduction in $N_\text{target}$, presenting a challenging trade-off. Our adaptive draft length prediction method achieves a better balance. Specifically, compared to EAGLE with a fixed draft length of 4, our method results in fewer wasted draft tokens and fewer target model forward passes, reducing both $T_\text{draft}$ and $T_\text{target}$, and ultimately enhancing overall throughput to 66.35 tok/s.

We also test the recently proposed acceptance rate modeling method DDD, which is based on the model-generated token probability. As shown in Table~\ref{tab:main-results} \methodname is comparable to DDD in the average acceleration performance, with a throughput of 64.44 tok/s and 64.43 tok/s respectively. DDD has more average accepted tokens, and this is possible because our model is relatively conservative in predicting the draft length, and DDD is more aggressive, generating more draft tokens and more waste tokens.

Our \methodname-DDD approach achieves an additional gain in decoding throughput, reaching an average of 65.20 tokens per second. \methodname-DDD also achieves the highest throughput on four out of six benchmarks. Notably, on the code generation task HumanEval, \methodname-DDD demonstrates the best acceleration performance. This is because we retain the longer of the two lengths, reducing the probability of premature draft exit and increasing the average acceptance length.

\subsection{Ablation Study}
In this section we explore the effectiveness of different designing choices in our method.

\subsubsection{Loss Penalty} 

\begin{figure}
    \centering
    \includegraphics[width=\linewidth]{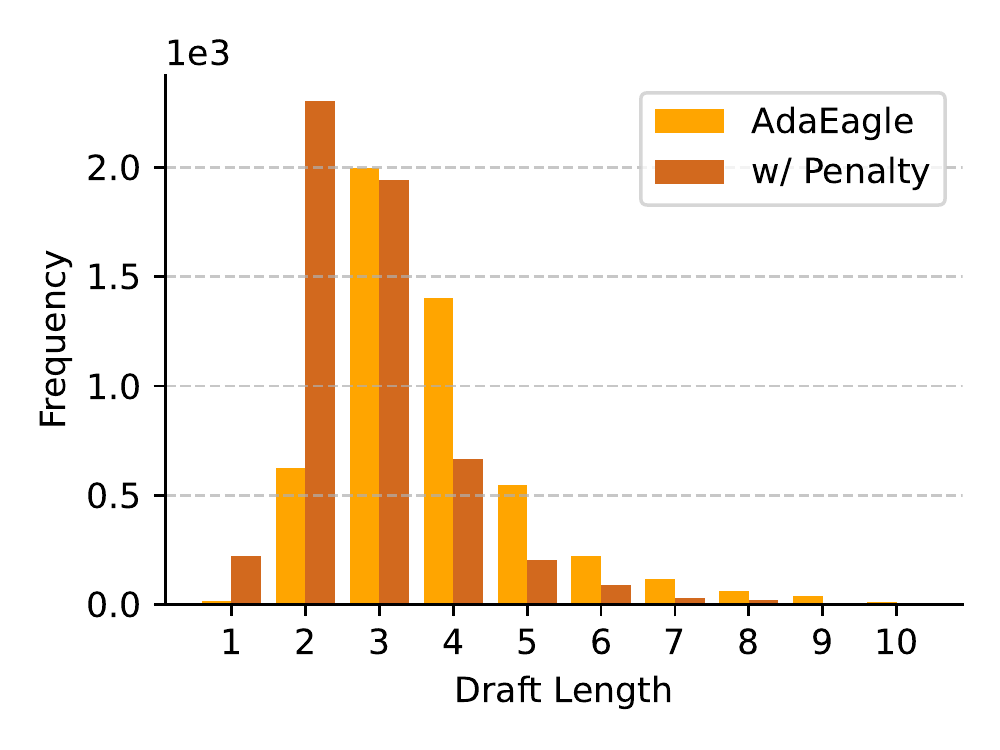}
    \caption{Comparison of draft length distributions with and without loss penalty.}
    \label{fig:dist_compare}
    \vspace{-1.5em}
\end{figure}

We train \methodname using L1 regression loss to predict the adaptive draft length logits. As discussed in Section~\ref{subsec:method_train}, shorter lengths are given greater tolerance compared to longer predictions. To address this, we apply a penalized L1 loss during training, which penalizes predictions that are shorter than the ground truth label. Table~\ref{tab:ablation-penalty} compares the impact of incorporating the penalized loss, showing that it improves both the average accepted length and throughput. Figure~\ref{fig:dist_compare} further illustrates the draft length distribution with and without the loss penalty, highlighting that the loss penalty encourages longer draft lengths.

\subsubsection{Regression v.s. Classification}
We also explore alternative approaches for draft length prediction, including regression and classification. In the regression approach, the model maps hidden states to a continuous score, whereas in the classification approach, it maps hidden states to a predefined set of classes. As shown in Table~\ref{tab:ablation-penalty}, regression outperforms classification. This is because regression can naturally model the ordinal relationships between different labels.

\subsubsection{\methodname-DDD thresholds}
Table~\ref{tab:threshold-results} presents the performance for various threshold values for the DDD component of \methodname-DDD, showing that the throughput remains stable and insensitive across different settings. Based on these results, we select a threshold value of -0.6 for the experiments reported in Table~\ref{tab:main-results}.

\begin{table}[t]
    \centering
    \small
    \begin{tabular}{lccccc}
        \toprule
        \textbf{Threshold} & -0.2 & -0.4 & -0.6 & -0.8 & -1.0 \\
        \midrule
        \textbf{Tok/s}     & 66.13 & 66.85 & 66.86 & 66.86 & 66.4 \\
        \bottomrule
    \end{tabular}
    \caption{Throughput results (Tok/s) for different threshold values. The thresholds correspond to the log probabilities, resulting in negative values. The ablation is conducted on the MT-Bench dataset.}
    \label{tab:threshold-results}
    \vspace{-2em}
\end{table}

\section{Related Works}
LLM is popular and capable. However, its AR inference is slow and costly. Significant methods for efficient LLM inference have been proposed~\cite{zhou_survey_2024}. This section focuses on the technique of speculative decoding (SD) and adaptive draft structure for SD. 

\paragraph{Speculative Decoding} Speculative decoding (SD) uses a draft-verify paradigm to realize lossless acceleration. SD frameworks can be roughly divided by the type of draft models~\cite{xia_SDsurvey_2024}: (1) Independent draft models. SpecDec~\cite{xia-2023-speculative} introduced a non-AR Transformer to generate multiple tokens simultaneously. Alternatively, many works~\cite{leviathan_fast_2023, chen_accelerating_2023, spector_accelerating_2023, Sun_SpecTr_2023} propose using smaller pre-trained models from the same LLM series (e.g., Llama2-7B and Llama2-70B~\cite{touvron_llama_2023}) to accelerate inference, which avoids extra training and preserves alignment in prediction behaviors due to shared tokenizers and training corpora. (2) Self-drafting eliminates the need for external draft models by leveraging the target LLM itself for drafting~\cite{Stern_Blockwise_2018, santilli-2023-accelerating, cai_medusa_2024, li_eagle_2024}. Recent techniques employ lightweight FFN heads to enable parallel~\cite{cai_medusa_2024} or AR~\cite{li_eagle_2024} token generation, reducing computational overhead. Other works~\cite{yang_predictive_2024, zhang_draft_2024} explore early exiting and layer skipping within the target model to improve drafting efficiency. However, most works use static draft structures, leaving the rewarding problem of adaptive draft generation unsolved.

\paragraph{Adaptive Draft Structure} 
An adaptive draft structure means that the draft model can adjust its generation structure (e.g., the length of the draft sequence, or the depth/width/shape of the draft tree) dynamically based on the speculation \textbf{context}.
Although some studies have recognized the need to make the draft structure dynamic, these methods still lack specialization and systematic approaches, as listed in Table \ref{tab:spec_dec_comparison}. 
Regarding the modeling approach of draft boundaries, some works use inherent approaches by leveraging the output distribution's confidence to determine the stopping point, as seen in~\cite{zhang_draft_2024, li_eagle-2_2024, liu_kangaroo_2024, brown_dynamicDDD_2024, wang_opt-tree_2024}, while others~\cite{huang_specdec_2024, mamou_dynamic_2024} employ extra modules such as additional trainable heads to determine whether to stop generating drafts. 

There are also some works orthogonal to the above, such as Sequoia~\cite{chen_sequoia_2024}, which is hardware-aware adaptive rather than context-aware adaptive. In summary, current models supporting adaptive draft structure lack performance, timeliness, and simplicity. The method proposed in this paper is the first SD framework that directly models the draft structure and achieves optimal performance based on EAGLE.

\section{Conclusion}
In this paper, we present a novel speculative decoding framework, \methodname, which is the first to implement adaptive draft control by explicitly modeling the optimal draft structure. 
By incorporating the Lightweight Draft Length Prophet (LDLP) module, \methodname predicts the optimal draft length to guide token generation, minimizing the number of forward passes for the large target model and maximizing overall decoding speed. 
Experimental results demonstrate that \methodname avxchieves decoding throughput closer to the oracle topline than existing SD methods. It provides a lossless speedup of 1.61\% over vanilla AR decoding, surpassing fixed-length SotA of 2\%, and shows generalizability and robustness across various test sets. 
Our approach opens the door for modeling adaptive draft structures in an explicit and dedicated ``end-to-end'' manner and can be integrated into a broad range of SD frameworks and sampling strategies.

\section*{Limitations}
Unlike EAGLE~\cite{li_eagle_2024} and EAGLE-2~\cite{li_eagle-2_2024} which use tree-based decoding, the method presented in this paper is currently limited to a sequential draft structure, so we only provide results under a greedy (top-1) decoding strategy. While a sequential structure allows us to clearly demonstrate our core idea with a single objective (length), tree-based structures require more complex and fine-grained control, such as the width and depth of each sub-tree, which we leave for future work.

Additionally, further exploration could be conducted on how LDLP can improve accuracy. Although this paper presents improvements based on EAGLE, the concept of directly modeling adaptive draft structures can, in fact, be applied to many other frameworks, such as~\cite{zhang_draft_2024}. Therefore, future research is needed to optimize LDLP, evaluate different SD frameworks, and examine the interactions between these elements.

\bibliography{custom}

\appendix

\section{Oracle-Guided Analysis of Optimal Draft Length}
\label{apd-sec:oracle}
Assume that when the input is token sequence $T_{a:b}$, output hidden states from the target model's second-to-top layer and the draft model's AR head are $F(T_{a:b}) = \{f(t_a), f(t_{j+1}) \dots, f(t_b)\}$ and $\hat{F}(T_{a:b}) = \{\hat{f}(t_a), \hat{f}(t_{j+1}) \dots, \hat{f}(t_b)\}$, respectively. Let the generated token sequence so far be denoted as $T_{1:j}$ and the draft token sequence based on $T_{1:j}$ be $\hat{T}_{j+1:j+k|j}$. Under the greedy decoding strategy, given the sequence history $T_{1:j}$, no matter what draft model outputs, the target model's output results are deterministic. We refer to the token sequence generated by the original target model (without drafts) as the \textit{formal} token sequence.

In SpecDec++~\cite{huang_specdec_2024}, the optimal draft length $k^*_j=\operatorname{optK}(j)$ is defined as the length such that minimizes the number of forward passes required by the target model. It can be inferred that $\hat{t}_{j+k^*_j|j}$ is accepted by the target model, while $\hat{t}_{j+k^*_j+1|j}$ is rejected. This occurs because in the SpecDec++ framework, generating a new draft token $\hat{t}_{j+i+1|j}$ depends only on the token history $T_{1:j}, \hat{T}_{j+1:j+i|j}$. Generating more than $k^*_j$ candidates results in unnecessary work by the draft model, while generating fewer than $k^*_j$ is sub-optimal as the rejected token $\hat{t}_{j+k^*_j+1|j}$ would still be generated in the next draft iteration, necessitating an additional forward pass from the target model.

However, in the EAGLE framework, draft generation depends not only on the token history but also on the hidden states produced by both the target and draft models, namely $F(T_{1:j})$ and $\hat{F}(\hat{T}_{j+1:j+i})$. If fewer than $k^*_j$, say $k'$ draft candidates are generated, the hidden history changes in the next draft iteration, meaning that a different draft token $\hat{t}_{j+k^*_j+1|j+k'+1}$ instead of the rejected $\hat{t}_{j+k^*_j+1|j}$ may be produced at the same position. That token, may equal $T_{j+k^*_j+1}$, and then be accepted by the target model, thus providing a chance of reducing the target model's forward passes. This implies that, even if we knew the next draft token could be accepted, generating it greedily is no longer guaranteed optimal.

To address this counter-intuitive situation, this paper introduces a reasonable assumption to resolve the dilemma.

\begin{assumption}
\label{asm:target-hist-longer}
    Let $j$ and $j'$ ($j' < j$) be the length of the generated formal token history, and currently the draft model is generating a draft token at position $m$ ($m > j$). Under EAGLE framework, we state that 
    \begin{equation}
        p_{\text{accept}}(\hat{t}_{m|j}) \ge p_{\text{accept}}(\hat{t}_{m|j'}).
    \end{equation}
\end{assumption}

To summarize, the more hidden states from the target, the more accurate guesses from the draft. This is reasonable because the draft hidden states and tokens are estimations of the target. Having a longer history from the target model allows the draft to generate output distributions that more closely align with those of the target model. Through this assumption, under EAGLE's framework, we obtain the following lemma:

\begin{lemma}
    \label{lma:max-len-best}
    Let the generated formal tokens be $T_{1:j}$. Under EAGLE framework, the optimal draft length is still $k^*_j=\operatorname{optK}(j)$ such that $\hat{t}_{j+k^*_j|j}$ is accepted by the target model, while $\hat{t}_{j+k^*_j+1|j}$ is rejected.
\end{lemma}

\paragraph{Prove}
\newcommand{\jA}{j_\textrm{A}}
\newcommand{\jAE}{j_\textrm{AEnd}}
\newcommand{\jB}{j_\textrm{B}}
It is obvious that a draft length larger than $k^*_j$ is still not optimal. 
Let $\jA=j+k^*_j+1, \jAE=\jA+\operatorname{optK}(\jA)$. In the next draft iteration, the draft model starts with $\hat{t}_{\jA+1|\jA}$, and ends with $\hat{t}_{\jAE|\jA}$, i.e., $p_\text{accept}(\hat{t}_{\jAE+1|\jA})=0$. Denote this draft scheme as scheme A.

When less than $k^*_j$ candidates, say $k'$ tokens are generated, let $\jB=j+k'+1$. Denote this draft scheme as scheme B. Since $\jB < \jA$, according to Assumption \ref{asm:target-hist-longer}, 
$$p_\text{accept}(\hat{t}_{\jAE+1|\jB}) \le p_\text{accept}(\hat{t}_{\jAE+1|\jA}) =0,$$
which means in the next draft iteration, the decoding progress for scheme B cannot exceed that of scheme A. Similarly, after the same number of draft iterations, scheme B is still left behind by scheme A. Equivalently, when generating the same number of formal tokens, scheme A needs fewer target model's forward passes than scheme B. \qed

Lemma \ref{lma:max-len-best} can also be easily extended to non-greedy decoding in terms of expectation.

\end{document}